%% file: main.tex
\documentclass[11pt]{article}

\usepackage{microtype}
\usepackage{graphicx}
\usepackage{subfigure}
\usepackage{booktabs} % for professional tables
\usepackage{adjustbox}

\usepackage{hyperref}
\usepackage{tcolorbox}
\usepackage{xspace}
\usepackage{multirow}
\usepackage{fancyvrb}

\newcommand{\llamalarge}{Llama-3.1 70B\xspace}

\newcommand{\OURS}{ExCoT\xspace}

\usepackage{times}

\usepackage[final]{acl}

\usepackage{amsmath}
\usepackage{amssymb}
\usepackage{mathtools}
\usepackage{amsthm}
\usepackage{tabularx}
\usepackage{lipsum}

\usepackage[capitalize,noabbrev]{cleveref}

\theoremstyle{plain}

\theoremstyle{definition}

\theoremstyle{remark}

\usepackage[textsize=tiny]{todonotes}

\title{\OURS: Optimizing Reasoning for Text-to-SQL with Execution Feedback}

\author{Bohan Zhai,~Canwen Xu,~Yuxiong He,~Zhewei Yao \\
Snowflake Inc. \\
\texttt{\{bohan.zhai, canwen.xu, yuxiong.he, zhewei.yao\}@snowflake.com}
}
\begin{document}
\maketitle

\input{abstract}

\input{introduction}

\label{submission}

\input{related}

\input{method_1.tex}

\input{results}

\input{conclusion}

\section*{Limitations}

Despite the value of chain-of-thought reasoning, our approach may still struggle under highly intricate schemas featuring complicated table relationships and domain-specific conventions. These specialized schemas frequently require deeper domain expertise or extensive training examples to ensure the model captures all relevant details. Further exploration of preference optimization with agents can mitigate these issues. Moreover, although exposing intermediate reasoning paths can foster greater transparency, it does not guarantee fully consistent or coherent logic at every step. Chain-of-thought traces may contain partial truths, redundant steps, or contradictions, which can complicate error diagnosis and reduce end-user confidence in the final SQL queries.

\section*{Broader Impact}

Our work enhances text-to-SQL automation, democratizing data access for non-experts in domains like healthcare or policy. However, errors in generated queries could propagate risks in critical applications; while execution feedback improves validity, users must verify outputs to avoid harmful decisions. Biases in training data or schemas may also perpetuate inequities if unchecked. Releasing models and data promotes transparency but requires safeguards against misuse (e.g., circumventing database security).

\bibliography{example_paper}

%%%%%%%%%%%%%%%%%%%%%%%%%%%%%%%%%%%%%%%%%%%%%%%%%%%%%%%%%%%%%%%%%%%%%%%%%%%%%%%
%%%%%%%%%%%%%%%%%%%%%%%%%%%%%%%%%%%%%%%%%%%%%%%%%%%%%%%%%%%%%%%%%%%%%%%%%%%%%%%
% APPENDIX
%%%%%%%%%%%%%%%%%%%%%%%%%%%%%%%%%%%%%%%%%%%%%%%%%%%%%%%%%%%%%%%%%%%%%%%%%%%%%%%
%%%%%%%%%%%%%%%%%%%%%%%%%%%%%%%%%%%%%%%%%%%%%%%%%%%%%%%%%%%%%%%%%%%%%%%%%%%%%%%
\newpage

\appendix
% \onecolumn
\input{appendix.tex}

%%%%%%%%%%%%%%%%%%%%%%%%%%%%%%%%%%%%%%%%%%%%%%%%%%%%%%%%%%%%%%%%%%%%%%%%%%%%%%%
%%%%%%%%%%%%%%%%%%%%%%%%%%%%%%%%%%%%%%%%%%%%%%%%%%%%%%%%%%%%%%%%%%%%%%%%%%%%%%%

\end{document}

%% file: abstract.tex
\begin{abstract}
Text-to-SQL demands precise reasoning to convert natural language questions into structured queries. 
While large language models (LLMs) excel in many reasoning tasks, their ability to leverage Chain-of-Thought (CoT) reasoning for text-to-SQL remains underexplored. 
We identify critical limitations: zero-shot CoT offers minimal gains, and Direct Preference Optimization (DPO) applied without CoT yields marginal improvements. 
We propose \OURS, a novel framework that iteratively optimizes open-source LLMs by combining CoT reasoning with off-policy and on-policy DPO, relying solely on execution accuracy as feedback. 
This approach eliminates the need for reward models or human-annotated preferences.
 Our experimental results demonstrate significant performance gains: \OURS improves execution accuracy on BIRD dev set from 57.37\% to 68.51\% and on Spider test set from 78.81\% to 86.59\% for LLaMA-3 70B, with Qwen-2.5-Coder demonstrating similar improvements. Our best model achieves state-of-the-art performance in the single-model setting on both BIRD and Spider datasets, notably achieving 68.53\% on the BIRD test set.\footnote{Our code is released at \href{https://github.com/snowflakedb/ArcticTraining/}{Arctic Training} and trained models are released at Huggingface: \href{https://huggingface.co/Snowflake/Qwen-2.5-coder-Arctic-ExCoT-32B}{Qwen-2.5-coder-Arctic-ExCoT-32B} and \href{https://huggingface.co/Snowflake/Llama-3.1-Arctic-ExCoT-70B}{Llama-3.1-Arctic-ExCoT-70B}}
\end{abstract}

%% file: introduction.tex
\section{Introduction}
\label{sec:introduction}

\begin{figure*}[t]  
    \centering  
    \includegraphics[width=\linewidth]{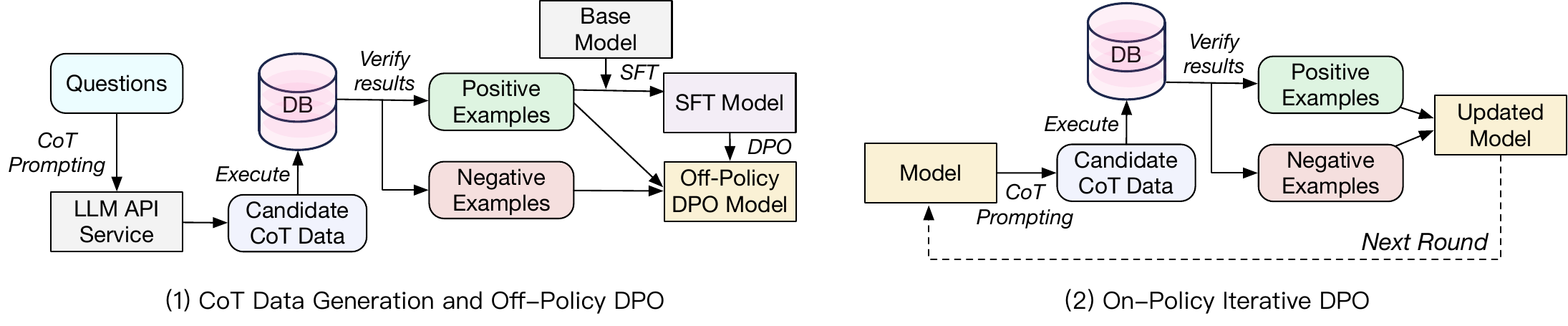}
    \caption{\textbf{The workflow of \OURS.} (1) We use a well-designed prompt to obtain candidate data from a LLM (GPT-4o is used in our experiment). We execute extracted SQLs on a local SQLite instance and compare the results with the ground truth. We use the positive examples to supervised fine-tune (SFT) the base model and construct the pairs for off-policy DPO. (2) We use the model trained with off-policy DPO to generate new candidate CoT data for on-policy DPO. We repeat this process iteratively for multiple rounds. }  
    \label{fig:workflow}  
\end{figure*} 

Text-to-SQL, the task of translating natural language questions into accurate SQL queries~\citep{li2024can, yu2018spider}, requires both domain knowledge and precise reasoning about query structures. This task is particularly challenging due to its reliance on schema linking (matching natural language entities to database columns), handling compositional syntax (e.g., nested queries, joins), and resolving ambiguities in user intent. In recent years, large language models (LLMs) have demonstrated remarkable capabilities in generating coherent, contextually rich responses with reasoning across a range of domains~\citep{o1,guan2025rstar, deepseekai2025deepseekr1incentivizingreasoningcapability}. These models are often trained with preference learning~\citep{ouyang2022training,rafailov2024direct} to promote the quality of their reasoning and accuracy on downstream tasks. However, it remains an open question how these LLMs can effectively leverage step-by-step reasoning, also referred to as Chain-of-Thought (CoT, \citealp{wei2022chain}), to improve performance in text-to-SQL, where errors in intermediate logic (e.g., misaligned joins or aggregation) often propagate irreversibly to final SQL outputs.

Prior work in text-to-SQL has explored specialized decoding strategies like execution-guided parsing~\citep{wang2021learning} or constrained inference~\citep{scholak2021picard}, but these methods typically focus on syntax correction rather than improving the model’s intrinsic reasoning. Meanwhile, preference optimization techniques like DPO have shown promise in aligning model outputs with human preferences in dialogue and summarization~\citep{rafailov2024direct}, yet their applicability to structured tasks like text-to-SQL remains understudied. 

Our investigations reveal three key observations that clarify the role of CoT in this setting: (1) zero-shot CoT --- without any dedicated training --- does not provides improvement over baseline methods in text-to-SQL tasks, likely because unstructured reasoning steps fail to address schema-specific dependencies or syntactic constraints; (2) directly applying Direct Preference Optimization (DPO) to text-to-SQL problem without incorporating a CoT does not substantially boost performance, as the absence of intermediate reasoning steps limits the feedback signal’s ability to pinpoint logical errors; and (3) when CoT is integrated, even a simple execution-based feedback signal can significantly enhance the model’s reasoning through DPO. We posit that the presence of intermediate reasoning steps in CoT exposes errors in logic or structure, thereby enabling the DPO process to correct those errors more effectively.

Building on these observations, we propose a novel framework, namely \textbf{Ex}ecution-Guided \textbf{C}hain-\textbf{o}f-\textbf{T}hought Direct Preference Optimization (\textbf{\OURS}), as illustrated in Figure~\ref{fig:workflow}. \OURS aligns open-source LLMs to text-to-SQL tasks by combining CoT with off-policy and on-policy iterative DPO, using only execution accuracy as a feedback mechanism. Crucially, this approach does not rely on a specialized reward model or human-annotated preference data; instead, it leverages the model’s own reasoning traces and downstream execution outcomes for iterative improvement. Notably, a similar conclusion in the domain of math and coding has also been reached by a concurrent research~\citep{deepseekai2025deepseekr1incentivizingreasoningcapability} that the answer correctness alone can be an effective signal. 
Our experiments demonstrate that the introduction of CoT into a DPO training loop allows the model to refine its reasoning steps, ultimately leading to more accurate SQL queries. 

In this paper, we make the following contributions:
(1) We demonstrate that zero-shot CoT does not offer gains in text-to-SQL, highlighting the necessity of explicit training and feedback mechanisms. Meanwhile, we show that applying DPO without CoT yields only marginal benefits, underscoring the importance of intermediate reasoning paths for effective alignment. (2) We introduce a simple yet powerful alignment strategy based solely on execution feedback (i.e., whether a generated query is correct), demonstrating that detailed CoT reasoning provides sufficient internal signals for the model to self-correct.

(3) We release models that achieve state-of-the-art performance in the single-model text-to-SQL setting. Our contributions include both the Llama-3.1 70B-based and Qwen-2.5-Coder 32B-based models, with our best model reaching 68.51\% and 68.53\% execution accuracy on the challenging BIRD benchmark dev set and test set, respectively.

Our work has broader implications for aligning LLMs to structured output tasks. By showing that execution feedback and self-generated reasoning traces suffice for iterative improvement, we reduce reliance on costly human annotations or reward models. This approach could generalize to other domains requiring precise structured generation, such as code synthesis or logical reasoning, where intermediate steps naturally admit validation (e.g., unit tests or formal verification).

%% file: related.tex
\section{Related Work}
The text-to-SQL task focuses on converting natural language queries into SQL commands to enable easy interaction with structured databases. Early work in this domain, such as Spider~\citep{yu2018spider}, established the complexity of the task by introducing benchmarks for cross-domain generalization and complex queries. Recent approaches to text-to-SQL can be broadly categorized into fine-tuning-based and prompting-based methods.

\paragraph{Fine-Tuning-Based Methods}
Fine-tuning pre-trained language models has been a dominant strategy in text-to-SQL research. RAT-SQL~\citep{wang2019rat} introduced a relation-aware schema encoding mechanism to enhance schema linking and query comprehension. PICARD~\citep{scholak2021picard} improved SQL query generation with constrained decoding, ensuring syntactic validity during generation. RESDSQL~\citep{li2023resdsql} decoupled schema linking and skeleton parsing for better modularity and adaptability. Furthermore, CodeS~\citep{CodeS} contributed open-source models that emphasized transparency and reproducibility in fine-tuned text-to-SQL systems.

\paragraph{Prompting-Based Methods}
Prompting large language models (LLMs) has gained attention due to its flexibility and minimal reliance on fine-tuning. ACT-SQL~\citep{zhang2023act} applied chain-of-thought reasoning in prompts to handle complex queries. \citet{gao2023text} evaluated LLMs on text-to-SQL tasks in zero-shot and few-shot settings, highlighting their effectiveness in reducing task-specific training. DIN-SQL~\citep{pourreza2024din} used a decomposed in-context learning approach, leveraging sub-query decomposition and self-correction. Similarly, CHESS~\citep{talaei2024chess} proposed contextual harnessing to streamline SQL synthesis. CHASE-SQL~\citep{pourreza2024chase} employed multi-agent modeling with divide-and-conquer, chain-of-thought reasoning, and instance-aware example generation. CHASE-SQL utilizes multiple prompts to generate SQL queries and refines them through a selection mechanism, optimizing for higher accuracy by selecting the best candidate from a set of generated SQL queries. 

Other works have tackled key challenges like schema linking, ambiguity resolution, and confidence calibration. Semantic matching in CRUSH4SQL~\citep{kothyari-etal-2023-crush4sql} addressed schema hallucination for improved linking accuracy. Reflexion~\citep{shinn2024reflexion} incorporated verbal reinforcement learning to iteratively refine SQL outputs. Calibration techniques, such as \citet{ramachandran2024texttosqlcalibrationneedask}, utilized model probabilities for better confidence scoring, while \citet{tian2023justaskcalibrationstrategies} explored strategies for eliciting calibrated confidence directly from LLMs. XiYan-SQL~\citep{gao2024xiyan} introduces a multi-generator ensemble framework that builds on similar concepts from CHASE-SQL. It combines multiple SQL generators and refines the output through an ensemble strategy, improving both the diversity and accuracy of generated SQL queries. XiYan-SQL uses a refiner to correct errors in the generated queries and a selection model to choose the best candidate from multiple generated SQL queries. The framework also integrates in-context learning and supervised fine-tuning to further enhance the quality of the SQL generation. 

Different from these works, \OURS unifies CoT and Direct Preference Optimization using only execution accuracy, eliminating the need for reward models or human annotation. This enables iterative, self-guided improvement of open-source LLMs, bridging CoT reasoning and preference alignment in text-to-SQL. While our work demonstrates strong single-model performance, \OURS is orthogonal to prompting-based frameworks and can be combined with those.

%% file: method_1.tex
\section{Execution-Guided Chain-of-Thought Preference Optimization} \label{sec:methodology}

\subsection{Overview} Text-to-SQL translation presents a unique challenge for large language models (LLMs), which must bridge the gap between complex natural language utterances and precise structured queries. Although chain-of-thought (CoT) reasoning has shown promise in many tasks, its effectiveness in text-to-SQL generation has not been fully explored. Moreover, while Direct Preference Optimization (DPO) can be used to align a model’s outputs with desired behaviors, naive applications of DPO without explicit reasoning often yield only marginal performance gains. To address these challenges, we introduce \textbf{\OURS}, an approach that seamlessly integrates CoT reasoning with DPO, using only execution-based correctness feedback (i.e., comparing the query’s result to the gold standard) to refine model performance.

Our method involves three key steps. First, we generate diverse chain-of-thought solutions for each text-to-SQL instance, ranging from simple to complex. Second, we apply an execution-based verification mechanism to label the correctness of each solution, thereby creating pools of correct (win) and incorrect (lose) responses. Finally, we perform preference optimization in both off-policy and on-policy settings. This multi-stage strategy reinforces correct solutions and demotes incorrect ones, progressively improving the model’s reasoning and query construction. By iterating these steps, we continuously refine the model without relying on human-annotated preferences or reward models. As we show later, combining CoT reasoning with DPO under execution feedback substantially advances text-to-SQL performance compared to zero-shot CoT or DPO alone.

\subsection{Chain-of-Thought Generation} \label{sec:cot_generation}

\paragraph{CoT Prompting}
Chain-of-Thought (CoT) reasoning provides a transparent path for LLMs to break down the query-construction process into intermediate steps. We explore three variants of CoT generation:
\textbf{(1) No-CoT (Direct SQL).} The model produces an SQL query directly from the input question and table schema, without revealing any intermediate reasoning. This baseline approach has the advantage of simplicity but often lacks interpretability and, as we show, fails to significantly leverage the model’s internal reasoning capabilities.
\textbf{(2) Simple-CoT.} Here, the model generates a short, high-level reasoning trace before outputting the final SQL. These concise rationales provide some insight into how the query is constructed but do not break down complex questions into sub-problems. Although improvements over No-CoT can be modest, Simple-CoT can serve as a stepping stone to more elaborate reasoning.
\textbf{(3) Complex-CoT (Divide-and-Conquer).} In complex real-world queries, a single chain of reasoning may not suffice to capture the multiple steps and sub-questions inherent in text-to-SQL. We therefore adopt a divide-and-conquer strategy akin to \citet{pourreza2024chase}, where the problem is decomposed into smaller sub-questions. Each sub-question is analyzed independently and answered with an intermediate SQL, which can in turn be referenced or combined to form a solution to the original query.

The exact prompts used in each type of prompting can be found in Appendix~\ref{appendix:prompts}.
Specifically, we begin with an analysis of the overall question and outline a pseudo-SQL template covering the required joins, filters, and aggregations. We then split the task into self-contained sub-questions, each associated with its own pseudo-SQL. These sub-queries are computed and combined in a later ``conquer'' step to produce a comprehensive SQL solution. Finally, we include a \emph{Simplification and Optimization Stage} where we merge or refine sub-queries to create a more efficient final SQL. This hierarchical structure helps the model handle the recursive nature of SQL operations, especially in queries involving nested sub-selects or multiple levels of aggregation.

\paragraph{CoT Data Synthesis} To generate reliable CoT exemplars, we employ few-shot prompting with GPT-4o~\cite{openai2024gpt4ocard}, which serves as our synthetic base model in this stage. We prepare prompts by randomly selecting examples from the BIRD and Spider training sets (9.2k and 8.6k examples, respectively) and providing three exemplars per prompt. Importantly, we supply the table schema and user question but exclude ground-truth SQL to encourage diverse and creative reasoning.

For each question, we generate multiple candidate solutions (up to 32), each potentially including a different intermediate reasoning chain. These output chains are then processed by our verification pipeline (detailed in Section~\ref{sec:off_policy_dpo}), which automatically executes the final SQL in a local database to judge correctness. To accommodate different reasoning styles and ensure robust coverage, we produce both short and complex CoTs.

After verification, only correct solutions --- i.e., those whose execution output matches the ground-truth query’s result --- are retained. This yields a high-quality Supervised Fine-Tuning (SFT) set, ultimately including 5.6k verified examples from BIRD and 6.1k from Spider. Generating multiple candidate CoTs per question boosts the chance of obtaining valid solutions, effectively making GPT-4o’s creativity work in our favor.

\subsection{Off-Policy DPO Alignment} \label{sec:off_policy_dpo}

Despite leveraging GPT-4o for diverse CoT solutions, not all generated SQL queries are correct. Simple heuristics such as discarding incorrect outputs risk under-utilizing the data. Instead, we adopt an \emph{off-policy} variant of Direct Preference Optimization \citep{pattnaik2024curry,xu2023contrastive}, where each question’s verified correct solutions (win) and its incorrect solutions (lose) form explicit preference pairs.

\paragraph{Execution-Based Feedback} To label solutions as correct or incorrect, we embed each relevant database schema in a local SQLite instance. We parse the final SQL from the model’s output—by convention, this is the last code block in the chain-of-thought reasoning. We then compare the query’s execution result to the ground-truth query result. Solutions that yield the same result are marked as correct, reflecting functional equivalence even if the SQL syntax differs.

\paragraph{Preference Pair Construction} After tagging each solution as correct or incorrect, we arrange them into positive (win) and negative (lose) pools. Inspired by \citet{pattnaik2024curry} and \citet{xu2023contrastive}, we measure the edit distance between each correct and incorrect pair, selecting those with the largest discrepancies for our final off-policy DPO training set. This strategy offers a type of “curriculum” in which the model is exposed to maximally dissimilar solutions, making the preference signal more salient and reinforcing a clear boundary between correct and incorrect reasoning.

\paragraph{Learning with Off-Policy DPO} Using these preference pairs, we fine-tune the base model to align its outputs toward correct solutions. Concretely, DPO adjusts model parameters so that, for each pair of responses, the likelihood of generating the winning solution is higher than that of generating the losing one. By leveraging \emph{both} correct and incorrect examples, we preserve rich data on how queries can go wrong, thus guiding the model away from pitfalls in future generations.

\subsection{On-Policy Iterative DPO with Execution Feedback} \label{sec:on_policy_dpo}

Although off-policy DPO on GPT-4o–generated data yields tangible performance gains, these synthetic solutions do not perfectly match the evolving distribution of the model under training. To address this mismatch, we further refine the model via \emph{on-policy} DPO, where newly generated solutions from the evolving model are incorporated back into the training loop.

\paragraph{Multi-Round On-Policy Loop} Our iterative refinement proceeds as follows: \textbf{(1) Start with Off-Policy DPO.} We first train the model using off-policy preference pairs derived from GPT-4o’s outputs. This step produces an initial model that is better aligned toward correct solutions. \textbf{(2) On-Policy Generation and Verification.} We use the newly updated model to generate fresh solutions (including chain-of-thought reasoning) for each question. We parse and execute these queries in a local database to verify correctness. \textbf{(3) On-Policy DPO Training.} We form win–lose pairs based on execution feedback and conduct a new round of DPO fine-tuning, aligning the model’s preferences toward the correct solutions it has just produced. \textbf{(4) Iterate.} We repeat on-policy generation and training for multiple rounds (e.g., three or more). Each round shifts the model’s distribution of generated solutions, prompting it to discover new correct queries that were previously out of reach. 
Though conceptually simple, this cycle steadily increases the model’s execution accuracy, demonstrating that repeated exposure to newly verified solutions can effectively correct remaining errors.

\subsection{Sampling Strategy} \label{sec:sampling_strategy} We employ a dynamic strategy for selecting which pairs of responses to include in our DPO training: \textbf{(1) Off-Policy Setting.} We select pairs displaying \emph{maximum} edit distance between correct and incorrect SQL. Pilot experiments showed that when operating on synthetically generated solutions from GPT-4o, maximizing differences between positive and negative pairs more effectively highlights the model’s most glaring mistakes. \textbf{(2) On-Policy Setting.} In contrast, for on-policy refinement, we select pairs with \emph{smaller} edit distance. Intuitively, these pairs reflect subtle errors in the model’s own distribution; correcting them can lead to rapid incremental improvements. Our pilot study suggested that, once the model has been partially aligned, focusing on near-miss solutions helps it refine its understanding more efficiently. We include the results in Section~\ref{sec:experiments}.

Through this combination of chain-of-thought generation, execution-based verification, and off- and on-policy DPO, our \OURS framework provides a powerful self-guided learning paradigm. The model consistently advances its text-to-SQL generation quality by leveraging intermediate reasoning, identifying correct solutions through execution feedback, and solidifying these discoveries via preference optimization. Notably, our approach requires neither human-curated rankings nor explicitly learned reward models, confirming that execution accuracy can serve as an effective and scalable alignment signal for text-to-SQL tasks.

%% file: results.tex
\begin{table*}[t]
    \centering
    \small
    \begin{tabular}{llccccc}
    \toprule
      \multirow{2}{*}{Model} & 
      \multirow{2}{*}{CoT Type} &
      \multicolumn{2}{c}{BIRD} & \multicolumn{2}{c}{Spider} \\
      \cmidrule(r){3-4} \cmidrule(l){5-6}
        & & EX\% & Valid\% & EX\% & Valid\% \\
    \midrule
    \textit{Base Model} \\
    LLaMA-3.1 70B~\citep{dubey2024llama} & No & 57.37 & 92.83 & 78.81 & 98.74 \\
    LLaMA-3.1 70B & Simple & 52.61 & 94.72 & 77.83 & 99.44 \\
    Qwen-2.5-Coder 32B~\citep{hui2024qwen2} & No & 58.93 & 89.31 & 79.32 & 99.72 \\
    Qwen-2.5-Coder 32B & Simple & 54.11 & 94.13 & 77.83 & 99.44 \\
    \midrule
    \textit{Stage 1: SFT with GPT-4o data} \\
    LLaMA-3.1 70B & No & 62.03 & 94.72 & 83.00 & 98.65 \\
    LLaMA-3.1 70B & Simple & 58.54 & 91.60 & 80.76 & 99.58 \\
    LLaMA-3.1 70B & Complex & 58.14 & 91.20 & 81.42 & 99.02 \\
    Qwen-2.5-Coder 32B & Complex & 59.65 & 91.13 & 81.23 & 99.12 \\
    \midrule
    \textit{Stage 2: Off-Policy DPO} \\
    LLaMA-3.1 70B & Simple & 64.39 & 96.74 & 83.93 & 99.53 \\
    LLaMA-3.1 70B & Complex & 66.30 & \textbf{98.50} & 82.49 & 99.86 \\
    Qwen-2.5-Coder 32B & Complex & 66.23 & 98.04 & 83.98 & 
    \textbf{99.95} \\
    \midrule
    \textit{Stage 3: On-Policy Iterative DPO} \\
    LLaMA-3.1 70B & No & 59.97 & 96.74 & 81.00 & 98.14 \\
    LLaMA-3.1 70B & Simple & 65.12 & 97.20 & 84.26 & 99.77 \\
    LLaMA-3.1 70B & Complex & \textbf{68.51} & \textbf{98.50} & \textbf{86.59} & 99.91 \\
    Qwen-2.5-Coder 32B & Complex & 68.25 & 98.37 & 85.14 & \textbf{99.95} \\
    \midrule
    \textit{Baselines} \\
    XiYanSQL-QwenCoder 32B~\citep{gao2024xiyan} & No & 63.75 & 95.83 & 81.42 & 99.72 \\
    OpenAI GPT-4o~\citep{openai2024gpt4ocard} & No & 54.69 & 90.87 & 76.53 & 99.21\\
    OpenAI GPT-4o & Simple & 54.04 & 94.79 & 76.01 & 99.35\\
    Anthropic Claude 3.5-Sonnet~\citep{anthropic2024claude} & No & 50.13 & 89.05 & 69.91 & 99.07 \\
    OpenAI o1-mini~\citep{o1} & Built-in & 52.41 & 86.38 & 75.13 & 98.09 \\
    OpenAI o3-mini & Built-in & 53.72 & 97.00 & 72.80 & 95.99 & \\
    
    \bottomrule
    \end{tabular}
    \caption{\textbf{Experimental results of \OURS.} Execution accuracy (EX\%) and SQL validity (Valid\%) on BIRD (dev set) and Spider (test set). \textit{Base Model} refers to the checkpoint trained only on broad natural language tasks. \textit{Stage~1} (SFT) adds GPT-4o-generated data to the supervised fine-tuning. \textit{Stage~2} (Off-Policy DPO) refines the model using preference pairs from GPT-4o–generated CoTs. \textit{Stage~3} (On-Policy Iterative DPO) further improves the model using repeated rounds of self-generated CoTs and execution-based verification.}
    \label{tab:main}
\end{table*}

\section{Experiments}
\label{sec:experiments}

\subsection{Experimental Settings}
\label{sec:exp_settings}
We conduct our experiments on two prominent cross-domain text-to-SQL benchmarks: the BIRD dataset \citep{li2024can}, consisting of 9.2k training queries, and the Spider dataset \citep{yu2018spider}, containing 8.6k training queries. Following standard practice, we measure performance using two metrics: (1) \textbf{execution accuracy} (\textbf{EX\%}), which reflects how often a generated SQL query returns the same result as the ground truth, and (2) \textbf{validity} (\textbf{Valid\%}), indicating whether the generated query parses and executes successfully without errors.

\paragraph{Model Initialization and Additional Data}
We primarily initialize our models from \llamalarge-base, a 70B-parameter LLM. To investigate the generality of our approach, we also employ Qwen-2.5-Coder 32B (the base version) \citep{hui2024qwen2, qwen2}, another code-focused LLM with 32B parameters. Before specializing in SQL tasks, we expose each base model to a range of supervised natural language datasets.
This step improves the model’s general language understanding and helps it handle broader reasoning patterns, thus laying a stronger foundation for subsequent fine-tuning on text-to-SQL.

\paragraph{Supervised Fine-Tuning (SFT)}
We perform one round of SFT on both the text-to-SQL data (approximately 12k queries across BIRD and Spider using the execution-based verified CoT data from GPT-4o) and the aforementioned natural language datasets. All experiments are conducted on 64 NVIDIA H100 GPUs. We set the local batch size to 2 per GPU (i.e., a global batch size of 128), use a learning rate of $1 \times 10^{-5}$, and train for 2 epochs. Each SFT stage takes 448 GPU hours. 

\paragraph{Preference Optimization}
Subsequent to SFT, we carry out both \emph{off-policy} and \emph{on-policy} DPO (see Sections~\ref{sec:off_policy_dpo}--\ref{sec:on_policy_dpo}). Similar to SFT, we use 64 NVIDIA H100 GPUs with a local batch size of 1 per GPU (i.e., 64 globally) and a learning rate of $1 \times 10^{-6}$. We train for 2 epochs in each DPO round. We conduct 1 round of off-policy DPO, followed by 2 rounds of on-policy DPO. The total compute cost for DPO is 160 GPU hours.

\paragraph{Chain-of-Thought Generation}
For our \textbf{off-policy} preference optimization, we rely on GPT-4o to produce up to 32 chain-of-thought (CoT) solutions per query. We then execute these queries in a local SQLite environment to determine correctness. In the \textbf{on-policy} setting, the model under training itself generates new solutions (also up to 32 per query), which are again verified via execution. We parse each CoT to retrieve the \emph{final} SQL block for execution. Pairwise comparisons between correct (win) and incorrect (lose) solutions feed into DPO updates.

\begin{table}[t]
    \centering
    \small
    % \resizebox{\linewidth}{!}{
    \begin{tabular}{ccccc}
    \toprule
       \multirow{2}{*}{Variant} & \multicolumn{2}{c}{BIRD} & 
       \multicolumn{2}{c}{Spider} \\
       \cmidrule(r){2-3} \cmidrule(l){4-5}
       & EX\% & Valid\% & EX\% & Valid\% \\
       \midrule
        1$\times$Off + 2$\times$On & \textbf{68.51} & \textbf{98.50} & \textbf{86.59} & \textbf{99.91}\\
        \midrule
        2$\times$On-Policy & 66.30 & 98.96 & 84.16 & 99.77 \\
        3$\times$On-Policy & 67.08 & 98.37 & 86.21 & 99.81 \\
    \bottomrule
    \end{tabular}
    % }
    \caption{Effect of pure on-policy DPO vs. applying both off-policy and on-policy DPO. We compare our default strategy (1 round of off-policy DPO and 2 rounds of on-policy DPO) with the variants where we remove the off-policy DPO round, or replace it with another round of on-policy DPO round.}
    \label{tab:offpolicy}
\end{table}

\subsection{Experimental Results}
\label{sec:exp_results}
\paragraph{Main Results}
Table~\ref{tab:main} summarizes our main results on the BIRD and Spider benchmarks. Notably, for LLaMA-3.1 70B model~\citep{dubey2024llama}, our \OURS approach improves BIRD performance from 58.14\% to 68.51\% with both off-policy and on-policy DPO, representing a significant 10.37\% gain. Similarly, on Spider, our \OURS approach improves the SFT model's performance of 82.49\% to 86.59\%. On both datasets, our approach achieves state-of-the-art performance in the single-model setting. With Qwen-2.5-Coder 32B~\citep{hui2024qwen2} model as the base model, our \OURS approach outperforms recent work XiyanSQL~\citep{gao2024xiyan} which uses the same base model, demonstrating the effectiveness of our approach. Our best model even outperforms state-of-the-art proprietary models, including OpenAI o1-mini~\citep{jaech2024openai} and o3-mini, which are built with chain-of-thought ability and optimized for reasoning. Notably, ExCoT achieves 68.53\% on BIRD test set, the highest performance among single models, as shown in Table~\ref{tab:test}.

Additionally, a case study is provided in Appendix~\ref{sec:case_study}.

\paragraph{Effect of Zero-Shot Chain-of-Thought Prompting}
As shown in Table~\ref{tab:main}, we can observe that injecting a \emph{Simple CoT} prompt does not provide meaningful performance improvement for LLaMA-3.1, Qwen-2.5-Coder or GPT-4o, suggesting that short, informal reasoning traces may not sufficiently guide the model toward accurate SQL.

\paragraph{Effect of SFT with GPT-4o Data (Stage~1)}
Adding correct solutions generated by GPT-4o to our supervised training set yields modest improvements in execution accuracy. For instance, \llamalarge with no-CoT SFT increases from 57.37\% to 62.03\% on BIRD and from 78.81\% to 83\% on Spider. Therefore, incorporating additional synthetic data does enhance coverage of diverse query patterns. For complex CoT setting, the model can learn the format of CoT from the GPT-4o output, setting the stage for further gains through preference optimization.

\paragraph{Effect of Off-Policy DPO (Stage~2)}
Off-policy DPO provides a more pronounced jump in performance. \llamalarge with Complex CoT goes from 58.14\% in SFT to 66.30\% execution accuracy on BIRD, and from 81.42\% to 82.49\% on Spider. A similar trend appears for Qwen-2.5-Coder (66.23\% on BIRD and 83.98\% on Spider). These improvements underscore how leveraging both correct and incorrect solutions --- rather than just discarding mistakes --- helps the model internalize fine-grained distinctions between correct and incorrect SQL.

To verify the effectiveness of this stage, we conduct an ablation study where we remove the off-policy DPO round, or replace it with another round of on-policy DPO. As shown in Table~\ref{tab:offpolicy}, involving off-policy DPO achieves better performance, as it provides diverse data to prepare the model for further on-policy DPO rounds.

\begin{table}[t]
    \centering
    \small
    % \resizebox{\linewidth}{!}{
    \begin{tabular}{cccccc}
    \toprule
       \multirow{2}{*}{DPO Round} & \multirow{2}{*}{Strategy} & \multicolumn{2}{c}{BIRD} \\ 
       % \multicolumn{2}{c}{Spider} \\
       \cmidrule(r){3-4} \cmidrule(l){5-6}
       & & EX\% & Valid\% \\ % & EX\% & Valid\% \\
       \midrule
        \multirow{3}{*}{Off-Policy} & Random & 64.08 & 98.31 \\ %& 81.93 & 99.67 \\
        & Nearest & 61.99 & \textbf{98.70} \\ % & \textbf{83.74} & 99.77 \\
        & Furthest & \textbf{66.30} & 98.50 \\ % & 82.49 & \textbf{99.86} \\
        \midrule
        \multirow{3}{*}{On-Policy} & Random & 66.36 & 97.65 \\ %& \textbf{86.12} & 99.86\\
        & Nearest & \textbf{67.21} & 97.72 \\ % & 85.28 & \textbf{99.91} \\
        & Furthest & 66.95 & \textbf{98.31} \\ %& 85.93 & 99.81 \\
    \bottomrule
    \end{tabular}
    % }
    \caption{Effect of different sampling strategies (curriculums) for off-policy and on-policy DPO (1 round).}
    \label{tab:curriculum}
\end{table}

\paragraph{Effect of On-Policy Iterative DPO (Stage~3)}
In the final stage, on-policy iterative updates yield the highest execution accuracy. With Complex CoT, \llamalarge reaches 68.51\% on BIRD and 86.59\% on Spider, while Qwen-2.5-Coder achieves 68.25\% on BIRD and 85.14\% on Spider. We note that even No-CoT variants see minor gains, but the biggest improvements occur under Complex CoT --- reinforcing our core assertion that detailed reasoning chains synergize effectively with iterative preference alignment. Notably, the baseline that is trained with DPO on SQLs without the CoT reasoning path only outperforms the corresponding base model trivially by 2.6\% and 2.2\% on BIRD and Spider, respectively. This suggests lack of reasoning path prevents the model from learning meaningful feedback signals in preference optimization.

\paragraph{Validity of Generated SQL}
Across all methods, the vast majority of queries remain syntactically valid (i.e., \textgreater90\% valid). Indeed, off-policy and on-policy DPO push validity above 97\% for most configurations. This high ratio ensures that improvements in execution accuracy are not merely byproducts of more well-formed queries; rather, the models are increasingly converging on \emph{functionally correct} queries.

\paragraph{Effect of Sampling Strategies for Off-Policy and On-Policy DPO} As a pilot study to understand the best curriculum setting, we train 1 round of off-policy and on-policy DPO with different sampling strategies, as described in Section~\ref{sec:sampling_strategy}. Based on the results shown in Table~\ref{tab:curriculum}, we choose the furthest and nearest strategies for off-policy and on-policy DPO, respectively.

\begin{figure}[t]
    \centering
    \includegraphics[width=\linewidth]{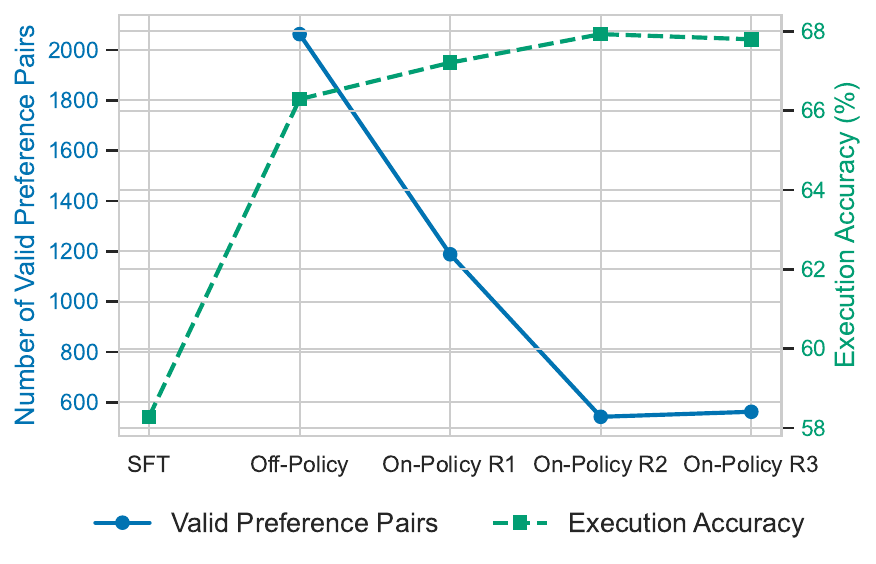}
    \caption{\textbf{Number of valid data pairs and corresponding execution accuracy on BIRD across successive training stages.} Although the pool of valid preference pairs decreases after off-policy DPO, each additional on-policy round of iterative DPO continues to boost execution accuracy, demonstrating that smaller yet targeted sets of self-generated examples effectively refine the model’s reasoning and SQL generation capabilities.}
    \label{fig:valid_pairs_vs_accuracy}
\end{figure}

\paragraph{Diminishing Effect of More DPO Rounds} We also add another round of on-policy DPO to explore if it can further improve the LLaMA model's performance. As shown in Figure~\ref{fig:valid_pairs_vs_accuracy}, the pool of valid preference pairs decrease significantly with more DPO rounds (from over 2,000 in the SFT stage to around 560 in the third on-policy round), indicating the exhaustion of contrastive pairs. The performance gains also exhibit a diminishing trend: while early on-policy rounds can lift accuracy substantially, later iterations yield narrower improvements or even degrade the performance. This indicates that while iterative refinement is crucial for polishing the model’s reasoning and SQL generation, each additional round contributes less benefit --- suggesting an eventual plateau in performance with repeated rounds.

\begin{figure}[t]
    \centering
    \includegraphics[width=\linewidth]{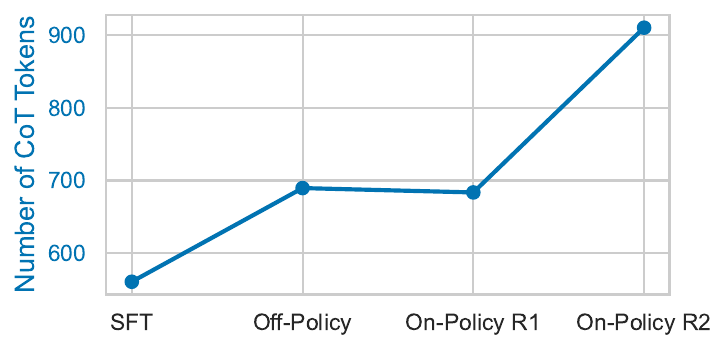}
    \caption{Number of CoT tokens across different training stages.}
    \label{fig:cot_len}
\end{figure}

\paragraph{Chain-of-Thought Length in Different Stages}
As shown in Figure~\ref{fig:cot_len}, we track how the average chain-of-thought (CoT) length (measured by the total number of tokens in the reasoning portion of each output) evolves across various stages of training. The CoT length increases from 560 tokens at the initial SFT stage to 910 tokens by the final on-policy round. This trend suggests the \OURS process not only boosts the model’s accuracy but also encourages it to elaborate further on intermediate reasoning steps.

\begin{table}[t]
    \centering
    \small
    % \resizebox{\linewidth}{!}{
    \begin{tabular}{ccc}
    \toprule
       \multirow{2}{*}{Model} & \multicolumn{2}{c}{BIRD} \\
        \cmidrule(l){2-3}
       & Dev EX\% & Test Ex\% \\
       \midrule
        \textbf{ExCoT-32B} & 68.32 & 68.19 \\
        \textbf{ExCoT-70B} & 68.51 & 68.53 \\
        SFT CodeS-15B & 58.47 & 60.37 \\
        SFT CodeS-7B & 57.17 & 59.25 \\
        DeepSeek 236B & 56.13 & 56.68 \\
        Mistral 123B & 53.52 & 55.84 \\
        GPT-4 & 46.35 & 54.89 \\
    \bottomrule
    \end{tabular}
    % }
    \caption{Results on BIRD test set comparing against other single models.}
    \label{tab:test}
\end{table}

%% file: conclusion.tex
\section{Conclusion and Future Work}

In this paper, we show that combining chain-of-thought (CoT) reasoning with direct preference optimization (DPO) can substantially boost text-to-SQL performance. Our \OURS framework relies solely on execution accuracy to iteratively refine open-source LLMs, eliminating the need for reward models or human annotations. The resulting gains on both BIRD and Spider demonstrate that aligning CoT with DPO offers a powerful path toward more accurate and self-guided text-to-SQL generation. Combined with SFT, \OURS improves execution accuracy on BIRD from 57.37\% to 68.51\% and on Spider from 78.81\% to 86.59\% for LLaMA-3.1 70B while also proven to be effective on another LLM, Qwen-2.5-Coder.

Concurrent works~\cite{deepseekai2025deepseekr1incentivizingreasoningcapability,team2025kimi} show that a free-style CoT process with the simple execution feedback can significantly improve model's performance tasks. 
We constraint our CoT template in this work, and we leave this free-style format as our feature work. 
Also, this work only applies offline-DPO to enhance model's text-to-SQL capability. 
More advanced approach, like Online-DPO~\citep{guo2024direct}, PPO~\citep{ouyang2022training}, and GRPO~\citep{shao2024deepseekmath}, are left for feature exploration.

%% file: appendix.tex
\appendix

\onecolumn
\section{Prompt Templates for CoT Data Generation}
\label{appendix:prompts}
\begin{table}[h!]
    \centering
    \begin{tabular}{|p{\linewidth}|}
        \hline
        \textbf{Non-Chain-of-Thought Prompt} \\
        \hline
        \textbf{System:} \\
        You are an AI assistant helping a data analyst write SQL queries to answer questions. \\
        \hline
        \textbf{User:} \\
        Below I will provide a DB schema and a question that can be answered by querying the provided DB. You will then write a SQL query enclosed in \texttt{```sql ...```} that answers the question (and nothing else). \\
        \textbf{Database Schema:} \{ Schema \} \\
        \textbf{Question:} \{ Question \} \\
        \hline
    \end{tabular} \\[0.25cm]
    
    \begin{tabular}{|p{\linewidth}|}
        \hline
        \textbf{Chain-of-Thought Prompt} \\
        \hline
        \textbf{System:} \\
        You are an AI assistant helping a data analyst write SQL queries to answer questions. \\
        \hline
        \textbf{User:} \\
        Below I will provide a DB schema and a question that can be answered by querying the provided DB. You will then write out your thought process in detail followed by a single SQL query enclosed in \texttt{```sql ...```} that answers the question. \\
        \textbf{Database Schema:} \{ Schema \} \\
        \textbf{Question:} \{ Question \} \\
        \hline
    \end{tabular} \\[0.25cm]
    
    \begin{tabular}{|p{\linewidth}|}
        \hline
        \textbf{Chain-of-Thought Prompt (Divide and Conquer)} \\
        \hline
        \textbf{System:} \\
        As a Text2SQL assistant, your main task is to formulate an SQL query in response to a given natural language inquiry. This process involves a chain-of-thought (CoT) approach, which includes a 'divide and conquer' strategy. \\
        In the 'divide' phase of this CoT process, we break down the presented question into smaller, more manageable sub-problems using pseudo-SQL queries. During the 'conquer' phase, we aggregate the solutions of these sub-problems to form the final response. \\
        Lastly, we refine the constructed query in the optimization step, eliminating any unnecessary clauses and conditions to ensure efficiency. \\
        \hline
        \textbf{User:} \\
        Below I will provide a DB schema and a question that can be answered by querying the provided DB. You will then write out your thought process in detail followed by a single SQL query enclosed in \texttt{```sql ...```} that answers the question. \\
        \textbf{Database Info:} \texttt{Database Schema:} \{ Schema \} \\
        \textbf{Question:} \texttt{Question:} \{ Question \} \\
        \textbf{Main Question:} \{ Main Question \} \quad \textbf{Analysis:} \{ Analysis \} \\
        \textbf{Pseudo SQL:} ```sql \{ Pseudo SQL \} ``` \\
        \textbf{Sub-questions:} \\
        \textbf{1.} \{ Sub-question \} \quad \textbf{Analysis:} \{ Analysis \} \quad \textbf{Pseudo SQL:} ```sql \{ Pseudo SQL \} ``` \\
        \textbf{2.} \{ Sub-question \} \quad \textbf{Analysis:} \{ Analysis \} \quad \textbf{Pseudo SQL:} ```sql \{ Pseudo SQL \} ``` \\
        \textbf{Final SQL Assembly:} ```sql \{ SQL \} ``` \\
        \textbf{Optimization:} \{ Analysis \} \quad ```sql \{ Optimized SQL \} ``` \\
        \hline
    \end{tabular}
    \caption{Chain-of-Thought prompts used in our CoT data generation.}
    \label{tab:cot_prompts}
\end{table}

\clearpage

\input{case_study}

%% file: case_study.tex
\twocolumn
\section{Case Study}
\label{sec:case_study}

We illustrate how our method progressively refines text-to-SQL answers on a very hard BIRD query by sampling each model 32 times to evaluate their success rate:

\begin{description}
\item[\textbf{Question.}]
For the school with the highest average score in Reading in the SAT test, what is its FRPM count for students aged 5-17?
\item[\textbf{Goal.}]
Identify the FRPM count for students aged 5-17 from the \texttt{frpm} table for the school with the highest average SAT Reading score from the \texttt{satscores} table.
\end{description}

\paragraph{Incremental Improvements Across Training Stages}
Table~\ref{tab:case_study} shows how many correct vs.\ incorrect solutions were generated at various stages for this particular example (BIRD index~10). Notably, GPT-4o failed to produce any correct solution in 32 trials for this query. After \textit{off-policy} DPO on GPT-4o–generated data, the model discovered a single correct SQL. In \textit{on-policy} rounds, where the model used its own generations as feedback, the number of correct solutions climbed further (5 at Round~1), demonstrating how iterative alignment via execution-based feedback unearths better queries.

\begin{table}[h]
\centering
\small
\begin{tabular}{lcc}
\toprule
\textbf{Stage}            & \textbf{\#Correct} & \textbf{\#Wrong} \\
\midrule
GPT-4o Generation        & 0                  & 32                \\
Off-Policy DPO           & 1                  & 31                \\
On-Policy DPO (Round 1)  & 5                  & 27                \\
On-Policy DPO (Round 2)  & 4                  & 28                \\
\bottomrule
\end{tabular}
\caption{\textbf{Outcome of different training stages on the BIRD example} (index 10). Each row shows how many correct vs.\ incorrect solutions were produced among 32 generated candidates.}
\label{tab:case_study}
\end{table}

\paragraph{Representative Chain-of-Thoughts and Final Queries}
Below, we highlight simplified excerpts of chain-of-thought (\textbf{CoT}) outputs from four snapshots: (1) GPT-4o, (2) Off-Policy DPO, (3) On-Policy DPO Round~1, and (4) On-Policy DPO Round~2. 
While each stage’s CoT differs in specific reasoning steps, they converge on similar final SQL templates that identify the school with the highest SAT Reading score and retrieve its FRPM count for students aged 5-17.

\paragraph{(1) GPT-4o (No Correct Solutions)}
\begin{quote}\small
\textit{CoT Excerpt:} 
``\dots We need to identify the school with the highest SAT Reading score, then join the \texttt{frpm} table to fetch the FRPM count. \dots''
\begin{verbatim}
WITH highest_reading AS (...)
SELECT FRPM_Count FROM frpm
WHERE CDSCode IN (SELECT cds
 FROM highest_reading);
\end{verbatim}
\end{quote}
Most GPT-4o attempts used complex subqueries or incorrect join conditions, resulting in 0/32 correct outputs despite the correct conceptual approach.

\paragraph{(2) Off-Policy DPO (1 Correct Solution)}
\begin{quote}\small
\textit{CoT Excerpt:} 
``\dots We can directly join \texttt{satscores} and \texttt{frpm} on the school code and order by \texttt{AvgScrRead} descending. \dots''
\begin{verbatim}
SELECT T2.`FRPM Count (Ages 5-17)`
FROM satscores AS T1
INNER JOIN frpm AS T2
  ON T1.cds = T2.CDSCode
ORDER BY T1.AvgScrRead DESC
LIMIT 1;
\end{verbatim}
\end{quote}
Off-policy DPO discovered a correct solution by prioritizing direct joins and ordering, ensuring a valid query within the SQLite environment.

\paragraph{(3) On-Policy DPO Round~1 (5 Correct Solutions)}
\begin{quote}\small
\textit{CoT Excerpt:}
``\dots To avoid errors, we directly join tables and order by reading scores. \dots''
\begin{verbatim}
SELECT T2.`FRPM Count (Ages 5-17)`
FROM frpm T2
JOIN satscores T1
  ON T1.cds = T2.CDSCode
ORDER BY T1.AvgScrRead DESC
LIMIT 1;
\end{verbatim}
\end{quote}
On-policy Round~1 produced multiple correct solutions by varying table aliases and consistently applying the correct ordering logic.

\paragraph{(4) On-Policy DPO Round~2 (4 Correct Solutions)}
\begin{quote}\small
\textit{CoT Excerpt:}
``\dots Simplify by using a single join query and ordering directly. \dots''
\begin{verbatim}
SELECT frpm.`FRPM Count (Ages 5-17)`
FROM frpm
JOIN satscores
  ON frpm.CDSCode = satscores.cds
ORDER BY satscores.AvgScrRead DESC
LIMIT 1;
\end{verbatim}
\end{quote}
In Round~2, the model maintained high accuracy with slight variations in table aliasing and ordering logic.

\paragraph{Summary of Observations} 
In this case, GPT-4o struggled to produce valid SQL due to unnecessary complexity. Our method:
\begin{enumerate}
\item \textbf{Off-Policy DPO} corrected GPT-4o’s approach by favoring direct joins and clear ordering.
\item \textbf{On-Policy DPO} consistently generated correct SQL by iteratively refining successful patterns.
\end{enumerate}